\documentclass[runningheads]{llncs}
\usepackage{graphicx}

\usepackage{tikz}
\usepackage{comment} 
\usepackage{amsmath,amssymb} 
\usepackage{color}

\newtheorem{defn}{Definition}
\newtheorem{thm}{Theorem}
\DeclareMathOperator*{\argmax}{arg\,max}
\makeatletter
\def\blfootnote{\xdef\@thefnmark{}\@footnotetext}
\makeatother

\begin{document}
\pagestyle{headings}
\mainmatter
\def\ECCVSubNumber{1044}  

\title{Ladybird: Quasi-Monte Carlo Sampling for \\ Deep Implicit Field Based 3D Reconstruction \\ with Symmetry} 

\titlerunning{Ladybird}
%
\author{Yifan $\text{Xu}^{\star}$\inst{1} \and
Tianqi $\text{Fan}^{\star}$\inst{2,3} \and
Yi $\text{Yuan}^{\dagger}$\inst{1}\and
Gurprit Singh\inst{3}}
\authorrunning{Y. Xu et al.}
%
\institute{Netease Fuxi AI Lab, Hangzhou, China \\
\email{\{xuyifan, yuanyi\}@corp.netease.com}\\
\and
Saarland Informatics Campus, Saarbruecken, Germany
\and
MPI for Informatics, Saarbruecken, Germany\\
\email{\{tfan,gsingh\}@mpi-inf.mpg.de}}
\maketitle

\begin{abstract}
Deep implicit field regression methods are effective for 3D reconstruction from single-view images. However, the impact of different sampling patterns on the reconstruction quality is not well-understood. In this work, we first study the effect of \emph{point set discrepancy} on the network training. Based on Farthest Point Sampling algorithm, we propose a sampling scheme that theoretically encourages better generalization performance, and results in fast convergence for SGD-based optimization algorithms. 
Secondly, based on the reflective symmetry of an object, we propose a feature fusion method that alleviates issues due to self-occlusions which makes it difficult to utilize local image features.
Our proposed system \emph{Ladybird} is able to create high quality 3D object reconstructions from a single input image. We evaluate Ladybird on a large scale 3D dataset (ShapeNet) demonstrating highly competitive results in terms of Chamfer distance, Earth Mover's distance and Intersection Over Union (IoU). 
\end{abstract}

\section{Introduction}


\blfootnote{$*$These two authors contribute equally. 
~~~~~~~~~~~$\dagger$Corresponding author.
\\ ORCIDs: Yi Yuan\orcidID{0000-0003-2507-8181} Gurprit  Singh\orcidID{0000-0003-0970-5835}~}

Due to the under-constrained nature of the problem, 3D object reconstruction from a single-view image has been a challenging task. Large shape and structure variations among objects make it difficult to define one dedicated parameterized model. Methods based on template deformation are often restricted by the initial topology of the template, and are not able to recover holes for instance.
Recently, deep learning based implicit fields regression methods have shown great potential in monocular 3D reconstruction. 
Mescheder et al.~\cite{mescheder2019occupancy} and DISN~\cite{xu2019disn}  create visually pleasing smooth shape reconstruction, with consistent normal and complex topology using implicit fields.

\begin{figure}[t!]
\centering
\includegraphics[height=9.0cm]{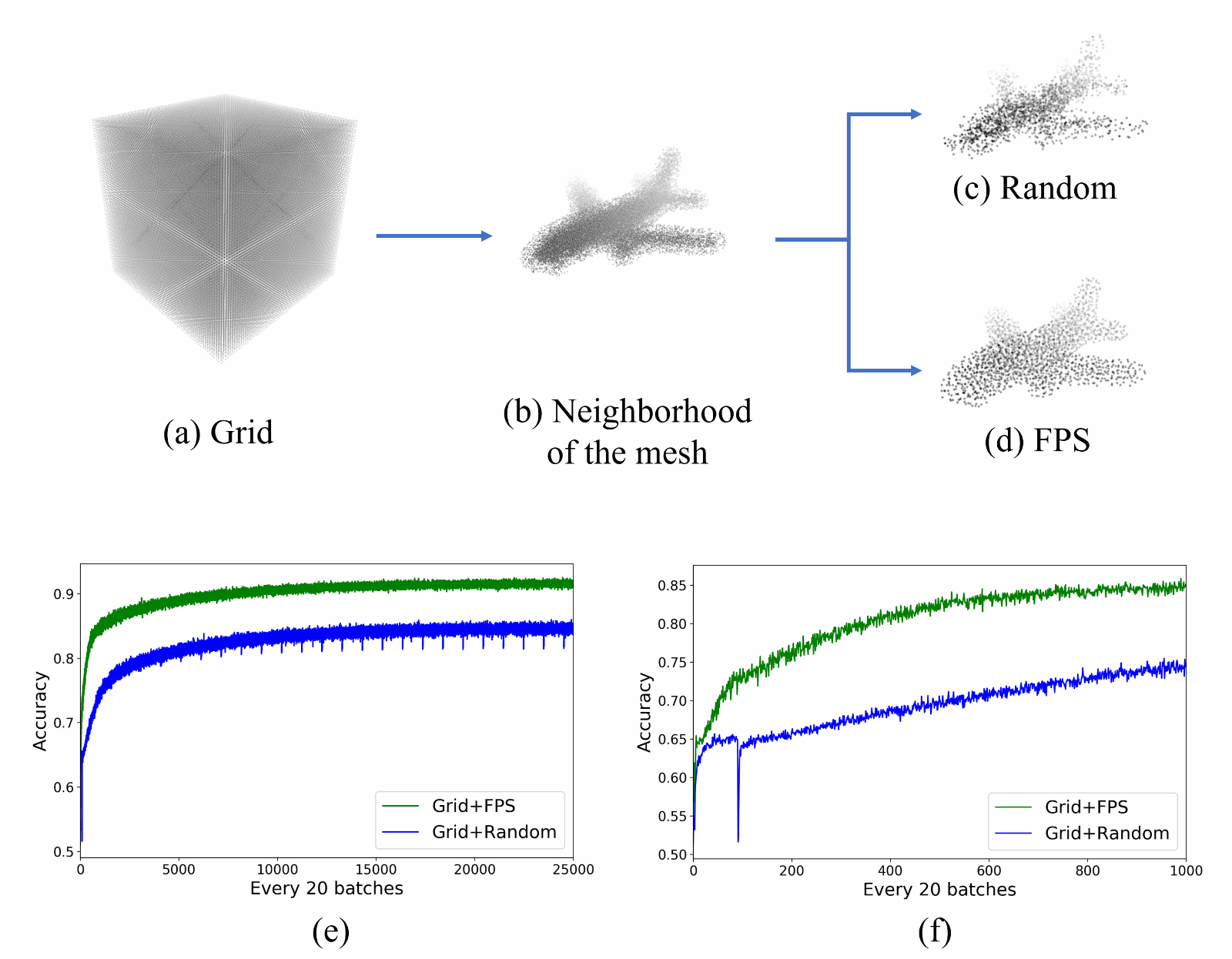}
\caption{Top: Demonstration of our sampling strategy for implicit field regression network training. A neighborhood of the mesh (b) is sampled from a set of dense grid points (a). A sparse set of points is sampled from (b) uniformly at random (c) or through FPS (d). Bottom: comparison of the training accuracy between Grid+FPS and Grid+Random sampling for the same network architecture during training. (e) is the plot of the training accuracy for the first 25 epoch. (f) is the plot of the training accuracy of the first epoch. Sampling with lower discrepancy  results in faster convergence and better accuracy during training.}
\label{fig:first}
\end{figure}

An implicit field is a real-valued function defined on $\mathbb{R}^3$ whose iso-surface recovers the mesh of interest. Common choices of implicit field are signed distance field, truncated signed distance field, or occupancy probability field. A network $g_w(I, p)$ is trained to predict the implicit field of point $p \in \mathbb{R}^3$, based on the input image $I$, where $w$ are the parameters which are optimized with stochastic gradient descent (SGD) type algorithms. This is followed by post-processing methods like marching cube and sphere tracing to reconstruct the mesh.

The loss function for the implicit field regression problem is the $L_2$ distance between the ground truth implicit field and the network $g_w$ predicted output. During training, a sparse set of 3D points need to be sampled in a compact region containing the mesh to approximate the optimization objective. 
We formulate this empirical loss as a Monte Carlo estimator.

While most prior discussion on sampling~\cite{mescheder2019occupancy} focuses on designing a probability measure for the integral that puts different weights for regions of different distance to the mesh surface, 
we look at the problem from a point view of discrepancy of the sample sets. 
When approximating an integral, different samplers have different error convergence rates with respect to the sample size~\cite{pilleboue2015variance} \cite{niederreiter1988low}.
Low discrepancy sequences/points or blue noise (in 2D) samples give better estimation, for instance, compared to random samples (white noise).

Given a set of locally uniform samples whose distance to the target mesh is bounded by a threshold, we show that farthest point sampling algorithm (FPS) can be used to select a sparse subset with low discrepancy for training $g_w$. An overview of our method is shown in Figure~\ref{fig:first}. Our proposed sampling scheme results in better generalization performance as it provides better approximation to the expected loss, thanks to the Koksma-Hlawka inequality \cite{kuipers2012uniform}. Empirically our sampling scheme also results in faster convergence for SGD-based optimization algorithms, which speeds up the training process significantly as shown in Figure~\ref{fig:first}(e,f).

\begin{figure}[t!]
\centering
\includegraphics[height=2.5cm]{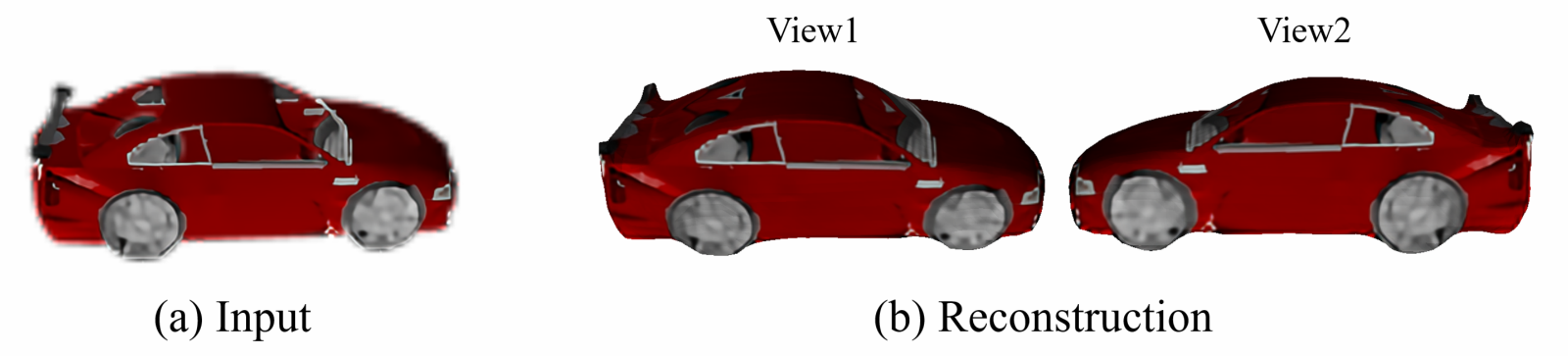}
\caption{Ladybird is able to produce high quality 3D reconstruction from a single input image. The consideration of symmetry allows recovering of occluded geometry and texture completion.}
\label{fig:big}
\end{figure}

Many deep 3D implicit field reconstruction works~\cite{mescheder2019occupancy} \cite{chen2019learning} explore the use of global shape encoding. While being good at capturing the general shape and obtaining interesting interpolation in the latent space, sometimes it is difficult to recover fine geometric details with only global features. Local features found via aligning image to mesh by modeling the camera are used to address the issue. However, for occluded points, it is ambiguous what local features should be used. Usually all the sampled points are projected to the images \cite{xu2019disn}, and hence points in the back use features of the points that occlude them.

As most man-made objects are symmetric about a plane, we observe that this problem can be alleviated via the consideration of reflective symmetry. For a symmetric pair of points $p$ and $q$, the implicit fields at $p$ and $q$ are the same, and often at least one of them is visible in the image. Hence we can use the local features of $q$ to improve the implicit field predication of $p$, which can also be understood as utilizing two-view information. Our feature fusion method imposes a symmetry prior on the network $g_w$, which gives significant improvement of the reconstruction quality as shown in Figure~\ref{fig:big}. Unlike previous works \cite{wang20193dn} \cite{yao2019front2back} that focus on the design of loss function, detection or encoding of symmetry, our method naturally integrates into the pixel-to-mesh alignment framework.

The advantage of spatially aligning the image to mesh and utilizing the corresponding local features is that the fine shape details and textures can be better recovered. However, when $p$ is occluded, the feature obtained by such alignment no longer has an intuitive meaning. Recently Front2Back \cite{yao2019front2back} addresses such issues by detecting reflective symmetries from the data and synthesizing the opposite orthographic view. Our approach is simpler and does not depend on symmetry detection.

\subsubsection{Mesh}

AtlasNet \cite{groueix2018atlasnet} represents a mesh as a locally parameterized surface and predicts the local patches from a latent shape representation learned via reconstruction objectives.
Mitchell et al. \cite{mitchell2019higher} proposes to represent 3D shapes using higher order functions.

Pixel2Mesh \cite{wang2018pixel2mesh} uses graph CNN to progressively deform an ellipsoid template mesh to fit the target. Features from different layers in the CNN are used to generate different resolution of details.
3DN \cite{wang20193dn} infers vertex offsets from a template mesh according to the image object's category, and proposes differentiable mesh sampling operator to compute the loss function.
SDM-NET \cite{gao2019sdm} uses VAE to generate a spatial arrangement of deformable parts of an object.
Pan et al. \cite{pan2019deep} proposes a progressive method that alternates between deforming the mesh and modifying the topology.
Mesh R-CNN \cite{gkioxari2019mesh} unifies object detection and shape reconstruction, with a mesh prediction branch that
first produces coarse cubified meshes which are refined with a graph convolution network.

DIB-R \cite{chen2019learning}, Soft Rasterizer \cite{liu2019soft} design differentiable rasterization layers that enable unsupervised training for reconstruction tasks. DIST \cite{liu2019dist} proposes an optimized differentiable sphere tracing layer for differentiable SDF rendering.

\subsubsection{Point Cloud and Voxel}
Fan et al. \cite{fan2017point} proposes a conditional shape sampler to predict multiple plausible point clouds from an input image.
Lin et al. \cite{lin2018learning} uses an auto-encoder to synthesize partial point clouds from multiple views, which is combined as a dense point cloud. Then the loss is computed via rendering the depth images from multiple views.
Li et al. \cite{li2018efficient} uses a CNN to predict multiple depth maps and corresponding deformation fields, which are fused to form the full 3D shape.

3D-R2N2 \cite{choy20163d} uses recurrent neural networks to generate voxelized 3D reconstruction.
Pixel2Vox \cite{xie2019pix2vox} uses encoder-decoder structures to generate a coarse 3D voxel.

\subsection{Sampling Methods in Monte Carlo Integration}
\begin{figure}
\centering
\includegraphics[height=4.5cm]{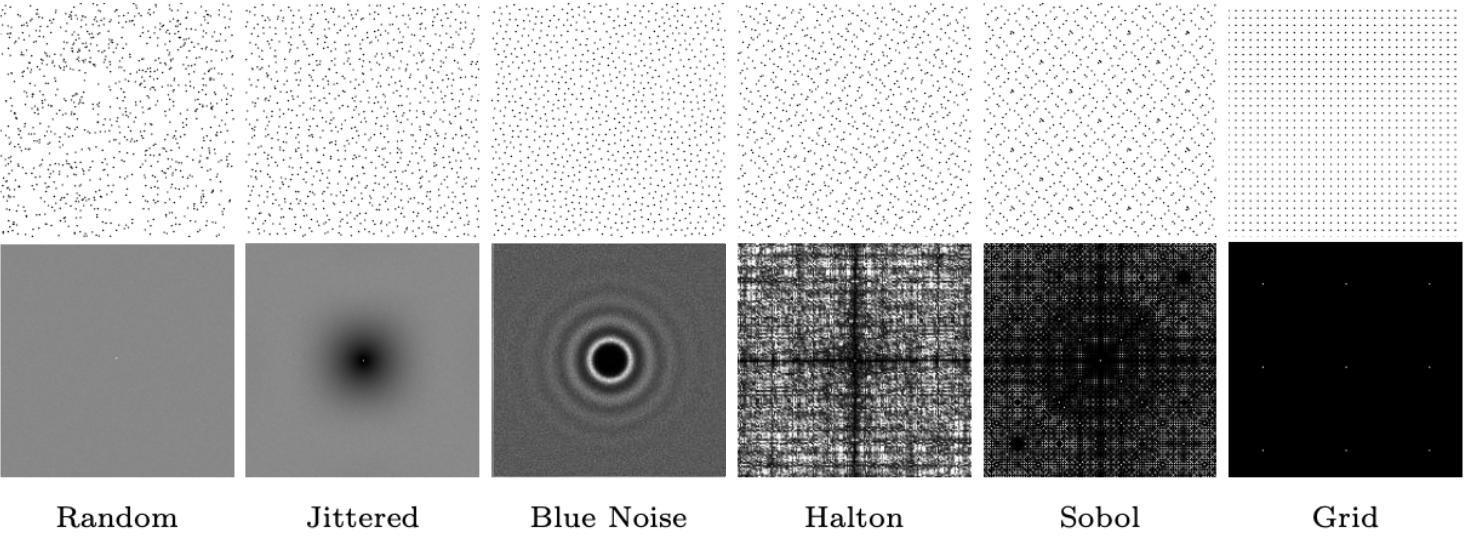}
\caption{Top row shows different point patterns for different samplers for $N=1024$ samples. Bottom row shows the corresponding \emph{expected} power spectra. Random samples are completely decorrelated which results in a flat power spectrum. 2D stratification (jittered) results in a power spectrum with small dark region around the center (DC frequency). For blue noise (Poisson Disk) sampler, this dark (no energy low-frequency) region is larger.  
However, for Halton and Sobol samplers, the corresponding power spectrum shows some spikes, but it preserves well the underlying stratification along dimensions which is characterized as a dark cross in the middle of the spectrum. Finally, a simple regular (Grid) pattern has a grid like power spectrum (zoom-in to the right-most bottom image to see the grid structure).}
\label{fig:samples}
\end{figure}

Realistic image synthesis involves evaluating very high-dimensional light transport integrals. (Quasi-)Monte Carlo (MC) numerical methods are traditionally employed to approximate these integrals which is highly error prone. This error directly depends on the sampling pattern used to estimate the 
underlying integral~\cite{singh19analysis}. 
These sampling patterns can be highly correlated. Fourier power spectra are commonly employed to characterize these  correlations among samples 
(Figure~\ref{fig:samples}). 

Blue noise samplers~\cite{singh19analysis} are well-known to show good improvements for low-dimensional integration problems whereas low-discrepancy~\cite{niederreiter1988low} samplers like Halton \cite{halton1964algorithm} and Sobol \cite{joe2008constructing} are more effective for higher dimensional problems. In this work, we use farthest point selection strategy \cite{eldar1997farthest} from any given pointset to select our samples.

\section{Our Approach}

\begin{figure}[t!]
\centering
\includegraphics[height=5.0cm]{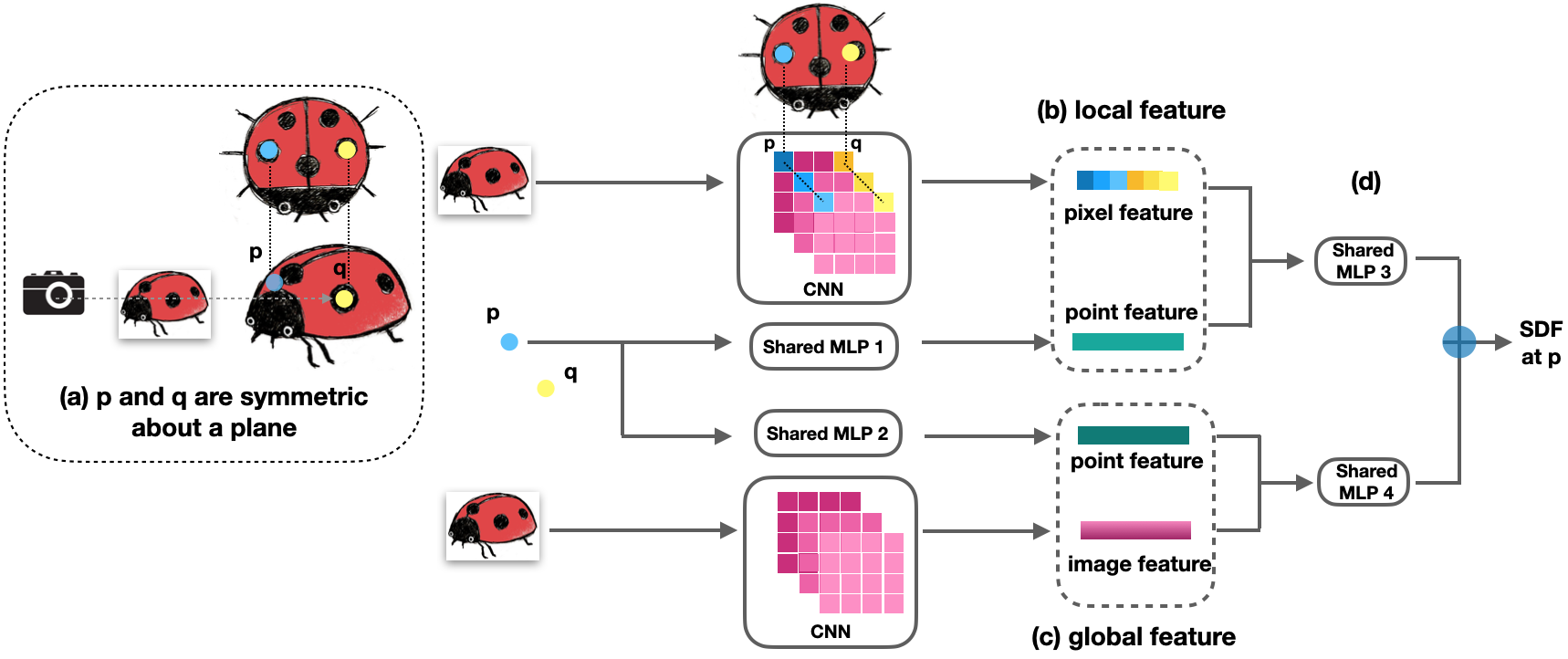}
\caption{Overview of Ladybird. a) $p, q \in \mathbb{R}$ are symmetric about a plane. Their projections to the image are found via a camera model. b) Local feature consists of point feature of $p$ and local image feature from pixels corresponding to $p$ and $q$. c) Global feature consists of point feature of $p$ and global image feature. d) Local feature and global feature are encoded through two MLPs whose parameters are shared among all $p \in \mathbb{R}$. In the end, marching cube is used to extract iso-surface.}
\label{fig:pipeline}
\end{figure}

We first start with a theoretical motivation for our sampling methods. This is followed by the proposed symmetric feature fusion module and our 3D reconstruction pipeline (illustrated in Figure~\ref{fig:pipeline}).

\subsection{Preliminary}
\label{subsec:prelims}
In Quasi-Monte Carlo integration literature, the equidistribution of a point set is tested by calculating the discrepancy of the set. This approach assigns a single quality number, the discrepancy, to every point set. The lower the discrepancy, the better (uniform) the underlying point set would be. 
We focus on the \emph{star discrepancy} of a point set, which computes discrepancy with respect to rectangular axis-aligned sub-regions with one of their corners fixed to the origin. Mathematically, the star discrepancy can be defined as follows:
\begin{defn}
Let $P = \{ x_1, x_2, ... , x_N\}$ be a set of points in $\mathbb{R}^d$, then the star discrepancy of $P$ is
\begin{equation}
D^*_N(P) = \sup_{B \in J} \| \frac{A(B; P)}{N} - \lambda_d(B) \|,
\end{equation}
where $\lambda_d$ is the Lebesgue measure on $\mathbb{R}^d$, $A(B; P)$ is the number of points in P that are in $B$, and $\displaystyle J = \{ \prod_{i = 1}^d [0, u_i) | 0 < u_i \leq 1 \}$.
\end{defn}
For a given point set or a sequence $P$ (stochastic or deterministic), the error due to sampling is directly related to the star discrepancy $D^*_N(P)$ of the point set $P$. 
This relation is given by the Koksma-Hlawka inequality \cite{kuipers2012uniform} as described below:
\begin{thm}
Let $I = [0,1]^d$ and $f$ is a function on $I$ with bounded variation $V(f)$. Then for any $x_1, x_2, ..., x_N \in I$,
\begin{equation}
\|\frac{1}{N} \sum_{i = 1}^N f(x_i) - \int_I f(x) dx\| \leq V(f) D^*_N( \{ x_1, ... , x_N \} )
\end{equation}
\end{thm}
The above inequality states that for $f$ with bounded variation, a point set with lower discrepancy gives less error when numerically integrating $f$. 

The distance between two implicit fields is an integral, and a set of points in needs to be sampled to approximate such integral which appears in the expected loss for deep implicit fields regression. By triangle inequality, using lower discrepancy sampler indicates a better bound on the generalisation error.

\subsection{Sampling}
\label{subsec:sampling}
Given an input image $I$, we denote  a neural network as $g_w(I, p)$ that predicts an implicit field of point $p \in \mathbb{R}^3$. 
Let $f_I(p)$ be the ground truth implicit field of the mesh $M$ from which $I$ is rendered, and let $S$ be the training set. To estimate the expected loss, we need to estimate the following:

\begin{equation}
\label{eq:eq1}
\sum_{I \in S} \int_{\mathbb{R}^3}(f_{I}(p)-g_w(I, p))^2 m(p) dp 
\end{equation}
where $m(p)$ is a probability density function in $\mathbb{R}^3$ supported in a compact region near the mesh $M$.

Instead of studying different choices for $m(p)$ and their effects on training, we study the impact of different sampling patterns on the integral estimation.

The error convergence rate of an estimator is greatly influenced by the sampling pattern~\cite{niederreiter1988low,pilleboue2015variance}. Sparse sampling could result in aliasing following the Nyquist-Shannon theorem. A better sampling strategy would allow faster convergence to the true integral resulting in better generalisation performance. Following the Koksma-Hlawka inequality, in order to better approximate the $L_2$ distance between $g_w$ and $f_{M}$---which indicates better generalisation of the network on different input points $p$---sample sets of lower discrepancy should be preferred.

In consideration of the time efficiency, usually we pre-compute the implicit field of a dense set of points around the mesh surface, where a sparse subset is chosen uniformly during training. Hence we consider the following problem: given a set of points $A$, how to select a subset $B \subseteq A$ consisting of $N$ points with low discrepancy. It is natural to consider farthest point sampling algorithm(FPS): initially $x_1 \in A$ is selected uniformly at random. Then iteratively,
\begin{equation}
x_i = \argmax_{x \in A \setminus B} (\min_{y \in B} d(x, y))
\end{equation}
is added to $B$. In Section~\ref{subsec:effect_samplers}, we show that compared to randomly selecting a sample subset $B$ from $A$, sampling using the FPS approach results in lower discrepancy.

\subsection{Feature Fusion Based on Symmetry}
\label{subsec:feature_fusion}
For a fixed camera model, let $\pi$ be the corresponding projection that maps 3D points to the image plane. Assume that the target mesh $M$ is symmetric about $xy$ plane
\footnote{ShapeNet data set is aligned, and most objects are symmetric about $xy$ plane.}, 
and $A$ is the rigid transformation such that the input image is formed via the composition $\pi \circ A$. In practice, either $A$ is known or $A$ is predicted via a camera network from input image.

For a point $p$ not too far from $M$, let $I_p$ be the pixel in the image that corresponds to $\pi(p)$. A convolution neural network (CNN) is used to extract features from the input image $I$. Let $F_p$ be the concatenation of feature vectors at $I_p$ in different layers of the CNN.

We can use $F_p$ to guide the regression of the implicit field at $p$. However when $p$ is occluded, the pixel value of $I_p$ is not determined by $p$ but by $r \in M$ with smallest z-buffer value whose projection $\pi(r)$ also lies in the pixel $I_p$. There is no clear relation between the implicit field at $p$ and that at $r$.

For a point $v = (x, y, z)$, such that $p = Av$, the symmetric point $q$ of $p$ is $A\bar{v}$ where $\bar{v} = (x, y, -z)$. The implicit field at $p$ should equal to that at $q$. Hence it is reasonable to include $F_q$ as part of the local feature of $p$, which we call feature fusion. One straight-forward and effective way to implement feature fusion is to concatenate $F_p$ and $F_q$.

\section{Experiments}
To show the effectiveness of our proposed system Ladybird, we provide quantitative as well qualitative comparisons to other methods. 
Our backbone network architecture is based on DISN~\cite{xu2019disn}. 
Our implementation of Ladybird is in Tensorflow 1.9 \cite{abadi2016tensorflow}, and the system is tested on Nvidia GTX 1080Ti with Cuda 9.0.
In all our experiments, Adam optimizer~\cite{Kingma2014} is used with $\beta = 0.5$ and an initial learning rate of 1e-4.

\subsection{Data Processing}
For dataset, we use ShapeNet Core v1 \cite{chang2015shapenet}, and use the official train/test split. There are 13 categories of objects. For each object, 24 views are rendered as in 3D-R2N2 \cite{choy20163d}. We randomly select 6000 images from the training set as the validation set, and our training set contains 726,600 images. The data is aligned and most objects (about 80 percent) are symmetric about $xy$ plane. We normalize the object mesh such that its center of mass is at the origin and the mesh lies in the unit sphere.

To efficiently and accurately compute the SDF values, we use polygon soup algorithm \cite{xu2014signed} to compute the SDF on $256^3$ grid points. After that, non-grid point SDF values are obtained through tri-linear interpolation.

For each mesh object, first we sample $256^3$ points $P_1$ using Grid, Jitter, or Sobol sampler \cite{UTK} and compute the corresponding SDF values. In Jitter, each grid point jitters with  Gaussian noise of mean 0 and standard deviation 0.02. We then sample a subset $P_2 \subset P_1$ consisting of 32,768 points from $P_1$ in the following way: from each SDF range $[-0.10, -0.03]$, $[-0.03, 0.00]$, $[0.00, 0.03]$, and $[0.03, 0.10]$, $1/4$-th of points are sampled uniformly at random.
During training time, a subset $P_3 \subset P_2$ consisting of 2048 points are sampled from $P_2$ uniformly at random or through FPS at each epoch. 
Depending on the sampling pattern used to sample $P_1$ (say $A$) and $P_3$ (say $B$), the resulting sampling pattern is denoted by $A+B$.

At test time, the SDF of $256^3$ grid points are predicated and marching cube is used to extract the iso-surface.

\subsection{Network Details}
\label{subsec:network_details}
We use a pre-trained camera pose estimation network from DISN \cite{xu2019disn}, to predict a rigid transformation matrix $A$ described in Section~\ref{subsec:feature_fusion}. VGG-16 is used as a CNN module to extract features form the input image. 
For a given point $p$, $F_p$ is the concatenation of features (Section~\ref{subsec:feature_fusion}) at different layers of VGG-16 at pixel $I_p$ that $p$ projects under the known or predicted camera intrinsics. Assuming $q$ and $p$ being symmetric about a plane,
the pixel feature of $p$ is one of the following:
\begin{enumerate}
\item Base: $F_p$ which is of dimension 1472.
\item Symm(Near): $F_p$ or $F_q$ depending on the one having smaller z-buffer value.
\item Symm(Avg): The average of $F_p$ and $F_q$.
\item Symm(Concat): The concatenation of $F_p$ and $F_q$.
\end{enumerate}
As shown in Figure~\ref{fig:pipeline}, the image feature is the output of VGG-16 (of dimension 1024). Two stream of point features are processed with two MLPs, each of parameters (64, 256, 512). Each stream is concatenated with pixel feature and image feature respectively, to form a local and a global feature. These global and local features are encoded through two MLPs, each of parameters (512, 256, 1), and the encoded values are added as the predicted SDF at $p$.

\subsection{Samplers Impact on Training}
\label{subsec:effect_samplers}
To assess the effect of different samplers on training, we set our pixel features to Base
(see Section~\ref{subsec:network_details}), use ground truth camera parameters and keep the batch size to 20.
\begin{table}[!t]
\begin{center}
\caption{Mean ($\times 0.01$) and standard deviation ($\times 0.01$) of star discrepancy of different samplers. A+B means we first sample $256^2$ points using sampling method A and then select a subset of size $n = 1024, 2048, 4096$ with method B.}
\label{table:star_discrepancy}
\begin{tabular}{c|c|c|c|c|c}
\hline
Sample size & Metric & Grid+Random & Grid+FPS& Jitter+FPS &Sobol+FPS\\
\hline
1024 & Mean & 4.51 & 3.86 & 6.49 & 4.35 \\
     & Std & 0.66 & 0.19 & 0.49 & 1.43\\
\hline
2048 & Mean & 2.98 & 2.48 & 6.07 & 2.51 \\
     & Std & 0.26 & 0.16 & 0.77& 0.47 \\
\hline
4096 & Mean & 2.34 & 1.96 & 6.10 & 1.65 \\
     & Std & 0.33 & 0.08 & 0.30 & 0.37\\
\hline
\end{tabular}
\end{center}
\end{table}
\begin{table}
\begin{center}
\caption{Effect of different samplers on the reconstruction results on ShapeNet test set. Method A+B means $P_1$ is sampled with A and $P_3$ is sampled with B. Metrics are class mean of CD ($\times 0.001$), and class mean of EMD ($\times 100$), computed on $2048$ points. Grid+FPS outperforms other methods.}
\label{table:sampling}
\begin{tabular}{c|c|c|c|c}
\hline
Metric & Grid+Random & Grid+FPS & Jitter+FPS &  Sobol+FPS \\
\hline
CD  & 10.17  & 8.43 & 19.88 & 11.33 \\
\hline
EMD & 2.71  & 2.57 & 2.92 & 2.84\\
\hline
\end{tabular}
\end{center}
\end{table}
In Table~\ref{table:star_discrepancy}, we report the star discrepancy of different samplers in 2D. We first sample $256^2$ points using Grid, Jitter, Sobol sampler in $[0,1]^2$, then selecting 1024, 2048, 4096 points uniformly at random or through FPS. In Jitter, each grid point jitters with Gaussian noise of mean 0 and standard deviation $0.01$. We  experimentally verify in 2D that Grid+FPS sampling has lower discrepancy and lower variance compared to Grid+Random.

\begin{figure}[!t]
\centering
\includegraphics[height=6.4cm]{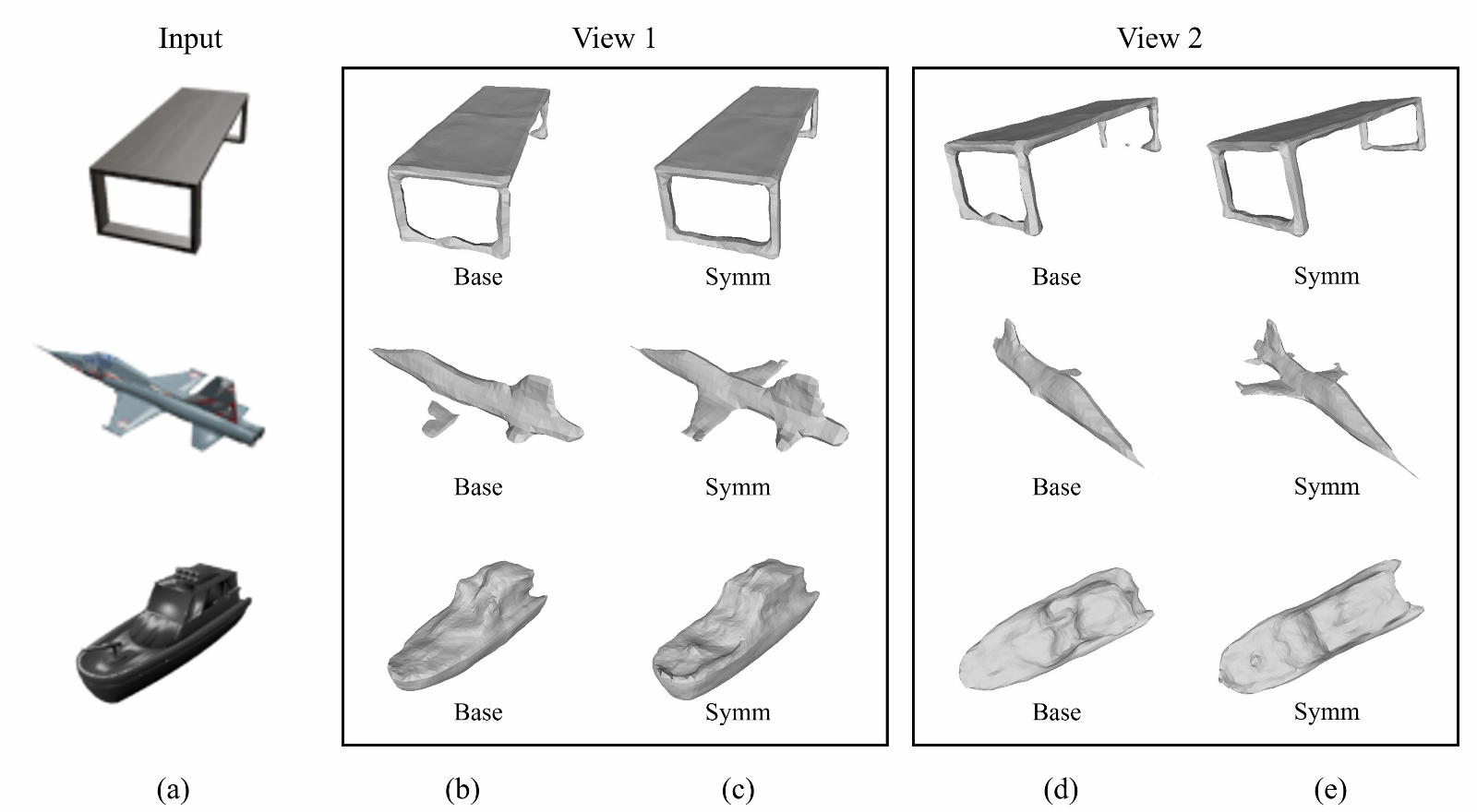}
\caption{Impact of feature fusion based on reflective symmetry. (a) indicates the input images. (b) and (d) are the reconstruction results using Base in two different views. (c) and (e) are the reconstruction results using Symm(Concat). We see that Symm(Concat) helps to improve the reconstruction quality.}
\label{fig:symm}
\end{figure}

\paragraph{Grid vs. Sobol:} 
The SDF validation accuracy of Sobol+FPS (0.914) is similar to that of Grid+FPS (0.917), which is higher than Grid+Random (0.825). However, SDF prediction is an intermediate step for the reconstruction task. Marching Cube is used to recover the mesh from the SDF, which requires SDF values at grid points. Due to this grid restriction imposed by Marching Cube, Grid sampling ensures better training/test data consistency. In addition, Grid+FPS and Grid+Random leads to more stable training results (cf. Sobol+FPS) due to lower std.
Our work advocates that Grid+FPS is suitable for 3D reconstruction based on deep implicit fields and marching cube.

In Table~\ref{table:sampling}, we report the comparison of reconstruction using different samplers in terms of Chamfer distance (CD)
\footnote{For two point set $S_1$ and $S_2$, CD is defined to be $\displaystyle \sum_{x \in S_1} \min_{y \in S_2} \|x-y\|^2_2 + \sum_{y \in S_2} \min_{ x \in S_1} \|x-y\|^2_2$.}
and Earth Mover's distance (EMD) \cite{villani2008optimal}. We see that Grid+FPS outperforms Grid+Random, Jitter+FPS, as well as  Sobol+FPS. Jitter+FPS performs the worst and its 2D analogue also has the highest star discrepancy. 
We observe that Grid+FPS reduces noisy phantom blocks around the mesh, and hence reduces the need for post-processing and cleaning. This property is highly desired, because sometimes the cleaning algorithm cannot distinguish between small components and noise. In addition, Grid+FPS encourages faster training convergence as shown in Figure~\ref{fig:first}.

\subsection{Effect of Feature Fusion Based on Symmetry}
To analyze the effect of symmetry-based feature fusion, we choose Grid+FPS sampling method. The corresponding batch size for this experiment is kept 16.

\begin{table}[!t]
\begin{center}
\caption{Comparison between different feature fusion operation evaluated on ShapeNet test set. Metrics are CD ($\times 0.001$), and EMD ($\times 100$), computed on $2048$ points. Groud truth camera parameters are used.}
\label{table:symmcompare}
\resizebox{\textwidth}{!}{
\begin{tabular}{c|c|ccccccccccccc|c}
Metric&  Local image feature  & plane & bench & box  & car & chair & display & lamp  &  speaker & rifle & sofa  & table & phone & boat  & Mean \\
\hline
CD
& Base
& 5.33  & 5.37  & 9.33  & 4.42  & 7.73  & 7.07    & 24.36 & 13.65   & \textbf{3.32}  & 5.78  & 9.37  & 8.13    & 5.79  & 8.43       \\
& Symm(Avg)
& 7.27  & 17.00  & 12.29  & 4.97  & 14.83  & 15.83 & 58.77 & 23.76  & 6.72  & 11.15 & 12.06 & 61.73 & 5.96 & 19.41 \\
& Symm(Near)
& 4.73  & 5.50  & 9.13  & 4.12  & 6.70  & 7.05 & 18.43 & 12.26  & 3.62  & 6.70 & 11.49 & 4.49 & 5.37 & 7.66 \\
& Symm(Concat)
& \textbf{3.86}  & \textbf{4.30}  & \textbf{8.04}  & \textbf{4.11}  & \textbf{5.43}  & \textbf{6.09}    & \textbf{14.10} & \textbf{10.53}   & 3.51  & \textbf{5.05}  & \textbf{8.13}  & \textbf{4.16}    & \textbf{4.92}  & \textbf{6.33}
\\
\hline
EMD
& Base
     & 2.35  & 2.30   & 2.91  & 2.47  & 2.66  & 2.44    & 4.21  & 3.19    & 1.69  & 2.29  & 2.78  & 1.95    & 2.14  & 2.57       \\
& Symm(Avg)
& 2.14  & 2.36  & 2.98  & 2.42  & 2.56  & 2.54 & 4.69 & 3.41  & 1.71  & 2.45 & 2.77 & 3.25 & 2.12 & 2.72 \\
& Symm(Near)
& 2.24  & 2.22  & 2.95  & 2.40  & 2.53  & 2.42 & 4.11 & 3.14  & \textbf{1.65}  & 2.38 & 2.85 & 1.89 & 2.11 & 2.53 \\
& Symm(Concat)
     & \textbf{2.07}  & \textbf{2.06}  & \textbf{2.80}  & \textbf{2.38}  & \textbf{2.32}  & \textbf{2.28}    & \textbf{3.59}  & \textbf{2.98}    & 1.73  & \textbf{2.18}  & \textbf{2.57}  & \textbf{1.85}    & \textbf{2.07}  & \textbf{2.38}     \\
\hline
IoU
& Base
     & 63.4  & 56.3   & 52.0  & 77.8  & 58.1  & 60.2    & 41.7  & 58.4    & 70.4  & 71.3  & 53.8  & 75.7    & 66.0  & 61.9       \\
& Symm(Near)
& 64.7  & 56.3  & 54.3  & 79.0  & 60.2  & 61.1 & 43.4 & 58.5  & 71.6  & 69.8 & 52.8 & 76.4 & 67.5 & 62.7 \\
& Symm(Concat)
     & \textbf{66.6}  & \textbf{60.3}   & \textbf{56.4}  & \textbf{80.2}  & \textbf{64.7}  & \textbf{63.7}    & \textbf{48.5}  & \textbf{61.5}    & \textbf{71.9}  & \textbf{73.5}  & \textbf{58.1} & \textbf{78.1}    & \textbf{68.8}  & \textbf{65.6}     \\
\hline
\end{tabular}
}
\end{center}
\end{table}

\begin{figure}
\centering
\includegraphics[height=2.3cm]{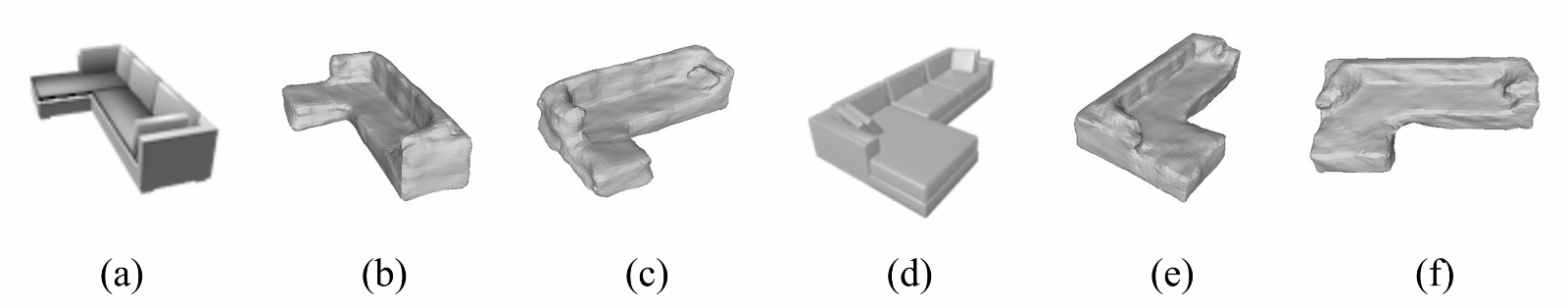}
\caption{Symm(Concat) can produce good reconstruction result for non-symmetrical object, without ground truth camera parameters. (a) and (d) are input images. (b) and (c) are reconstruction result of (a) rendered from 2 different views. (e) and (f) are reconstruction result of (d) from two different views.}
\label{fig:nonsymm}
\end{figure}

In Table~\ref{table:symmcompare} and Figure~\ref{fig:symm}, we compare the effects of different feature fusion operations that are defined in Section~\ref{subsec:network_details} on the reconstruction result from ShapeNet. Ablation study shows that Symm(Near) and Symm(Concat) improve the reconstruction results.
We see that concatenation of features from symmetrical pair performs the best. The reason is that Symm(Concat) better utilizes additional information comparing to Symm(Near) and Symm(Avg). When both $p$ and its symmetry point $q$ are visible in the image, the pixel features of $p$ and $q$ are both helpful for recovering the local shape at $p$. We observe that Symm(Concat) is able to produce reconstruction result for non-symmetrical object as shown in Figure~\ref{fig:nonsymm}. It has the interpretation of adding the most promising additional local feature based on a symmetry prior.

\begin{figure}
\centering
\includegraphics[height=12.5cm]{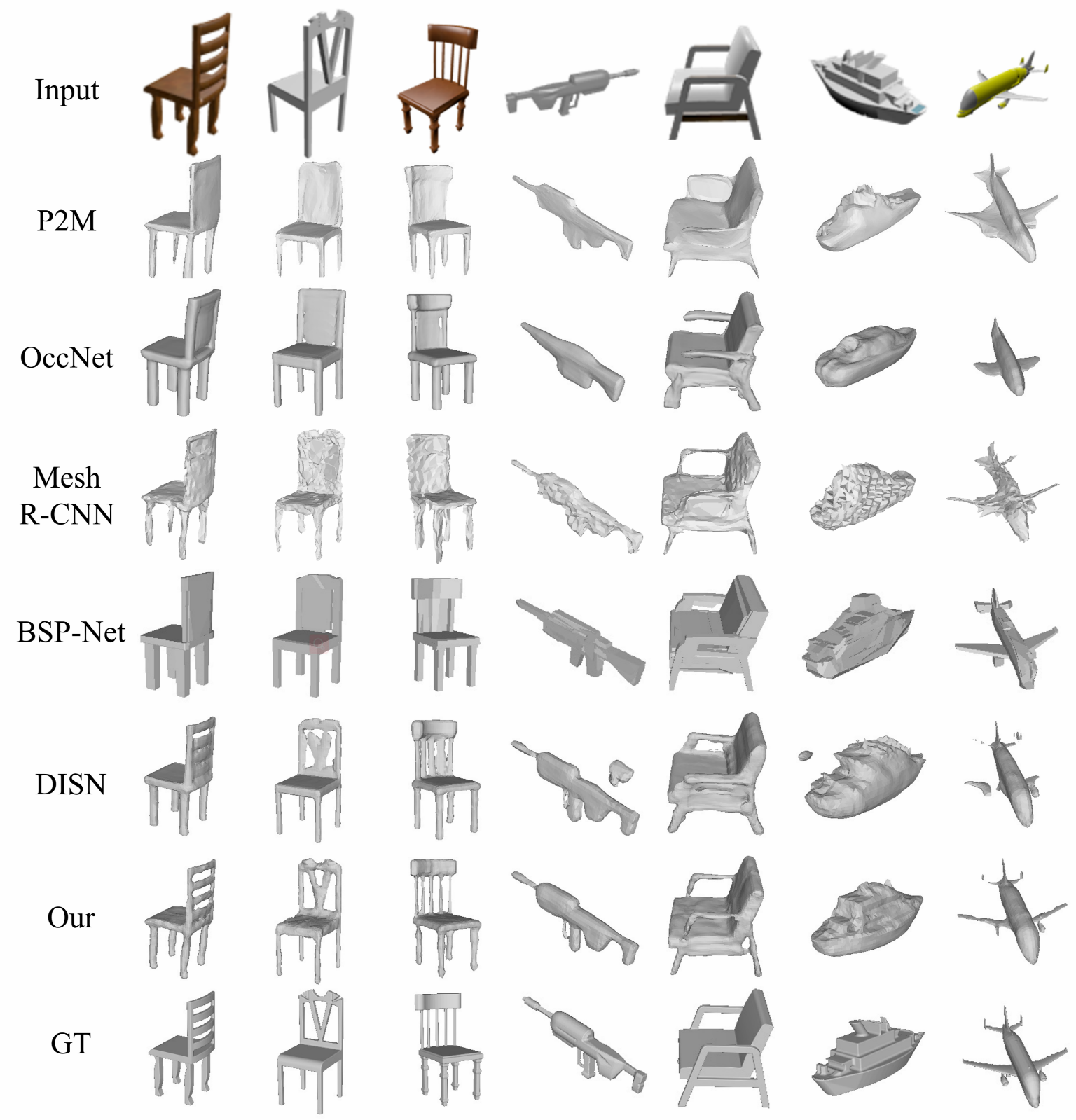}
\caption{Qualitative comparison with other methods. The first row contains the input image. The released model of P2M (Pixel2Mesh)\cite{wang2018pixel2mesh}, OccNet \cite{mescheder2019occupancy}, Mesh-R-CNN \cite{gkioxari2019mesh}, BSP-NET \cite{chen2020bsp}, DISN \cite{xu2019disn} are used to generate the results. The last row GT contains the ground truth meshes.}
\label{fig:comp}
\end{figure}

\subsection{Comparison with Other Methods}

In this subsection, the sampling method is Grid+FPS. The pixel feature is Symm(Concat). Camera parameters are estimated using the network mentioned in Section 4.2. 

We report comparison with other state-of-the-art methods in terms of CD, EMD, and IoU. From Table~\ref{table:totalcomp}, we see that Ladybird outperforms other methods. 
Figure~\ref{fig:comp} shows qualitative comparison of Ladybird with other methods. We see that Ladybird is able to reconstruct high quality mesh with fine geometric details from a single input image. Note that due to the difference between the train/test split of OccNet \cite{mescheder2019occupancy} and that of ours, we evaluate OccNet \cite{mescheder2019occupancy} on the intersection between two test sets.

\begin{table}
\begin{center}
\caption{Evaluations on ShapeNet Core test set for various methods. Metrics are CD ($\times 0.001$), EMD ($\times 100$) and IoU ($\%$, the larger the better), computed on $2048$ points. $Ours_{cam}$ is Ladybird with estimated camera parameters, and $Ours$ is Ladybird with ground truth camera parameters.}
\label{table:totalcomp}
\resizebox{\textwidth}{!}{
\begin{tabular}{c|c|ccccccccccccc|c}
Metric&  Method  & plane & bench & box  & car & chair & display & lamp  &  speaker & rifle & sofa  & table & phone & boat  & Mean \\
\hline
CD  & AtlasNet \cite{groueix2018atlasnet}
     & 5.98  & 6.98  & 13.76 & 17.04 & 13.21 & 7.18    & 38.21 & 15.96   & 4.59  & 8.29  & 18.08 & 6.35  & 15.85 & 13.19 \\
& Pixel2Mesh \cite{wang2018pixel2mesh}
     & 6.10  & 6.20  & 12.11 & 13.45 & 11.13 & \textbf{6.39}    & 31.41 & \textbf{14.52}   & 4.51  & \textbf{6.54}  & 15.61 & 6.04  & 12.66 & 11.28 \\
& 3DN \cite{wang20193dn}
     & 6.75  & 7.96  & \textbf{8.34}  & 7.09  & 17.53 & 8.35    & \textbf{12.79} & 17.28   & \textbf{3.26}  & 8.27  & 14.05 & 5.18  & 10.20 & 9.77  \\
& IMNET \cite{chen2019learning}
     & 12.65 & 15.10 & 11.39 & 8.86  & 11.27 & 13.77   & 63.84 & 21.83   & 8.73  &   10.30 & 17.82 & 7.06  & 13.25 & 16.61       \\
& 3DCNN \cite{xu2019disn}
& 10.47 & 10.94 & 10.40 & 5.26  & 11.15 & 11.78   & 35.97 & 17.97   & 6.80  & 9.76  & 13.35 & 6.30  &  9.80 & 12.30 \\
& OccNet \cite{mescheder2019occupancy}
& 7.70  & 6.43   & 9.36  & 5.26  & 7.67  & 7.54    & 26.46  & 17.30    & 4.86  & 6.72  & 10.57  & 7.17   & 9.09  & 9.70 \\
& DISN \cite{xu2019disn}
& 9.96    & 8.98    & 10.19 & 5.39 & 7.71   & 10.23 & 25.76 &      17.90 & 5.58 &     9.16 & 13.59 & 6.40 & 11.91 & 10.98  \\
& Ours$_{cam}$
    & \textbf{5.85}  & \textbf{6.12}   & 9.10  & \textbf{5.13}  & \textbf{7.08}  & 8.23    & 21.46  & 14.75    & 5.53  & 6.78  & \textbf{9.97}  & \textbf{5.06}    & \textbf{6.71}  & \textbf{8.60} \\
& Ours
& 3.86  & 4.30  & 8.04  & 4.11  & 5.43  & 6.09    & 14.10 & 10.53   & 3.51  & 5.05  & 8.13  & 4.16    & 4.92  & 6.33
\\
\hline
EMD & AtlasNet \cite{groueix2018atlasnet}
     & 3.39  & 3.22  & 3.36  & 3.72  & 3.86  & 3.12    & 5.29  & 3.75    & 3.35  & 3.14  & 3.98  & 3.19  & 4.39  & 3.67  \\
& Pixel2Mesh \cite{wang2018pixel2mesh}
     & 2.98  & 2.58  & 3.44  & 3.43  & 3.52  & 2.92    & 5.15  & 3.56    & 3.04  & 2.70  & 3.52  & 2.66  & 3.94  & 3.34  \\
& 3DN \cite{wang20193dn}
     & 3.30  & 2.98  & 3.21  & 3.28  & 4.45  & 3.91    & 3.99  & 4.47    & 2.78  & 3.31  & 3.94  & 2.70  & 3.92  & 3.56  \\
& IMNET\cite{chen2019learning}
     & 2.90  & 2.80  & 3.14  & 2.73  & 3.01  & 2.81    & 5.85  & 3.80    & 2.65  &   2.71  & 3.39  & 2.14  & 2.75  & 3.13       \\
& 3DCNN  \cite{xu2019disn}
     & 3.36  & 2.90  & 3.06  & \textbf{2.52}  & 3.01  & 2.85    & 4.73  & 3.35    & 2.71  & 2.60  & 3.09  & 2.10  & 2.67  & 3.00 \\
& OccNet \cite{mescheder2019occupancy}
& 2.75  & 2.43   & 3.05  & 2.56  & 2.70  & \textbf{2.58}    & \textbf{3.96}  & 3.46    & 2.27  & \textbf{2.35}  & 2.83  & 2.27    & 2.57  & 2.75 \\
& DISN \cite{xu2019disn}
     & 2.67     & 2.48 &   3.04  & 2.67 &  2.67  & 2.73 & 4.38 & 3.47 & 2.30 & 2.62 & 3.11 & 2.06 & 2.77 & 2.84       \\
& Ours$_{cam}$
& \textbf{2.48}  & \textbf{2.29}   & \textbf{3.03}  & 2.65  & \textbf{2.60}  & 2.61    & 4.20  & \textbf{3.32}    & \textbf{2.22}  & 2.42  & \textbf{2.82}  & \textbf{2.06}    & \textbf{2.46}  & \textbf{2.71} \\
& Ours
     & 2.07  & 2.06  & 2.80  & 2.38  & 2.32  & 2.28    & 3.59  & 2.98    & 1.73  & 2.18  & 2.57  & 1.85    & 2.07  & 2.38     \\
\hline
IoU
& AtlasNet \cite{groueix2018atlasnet}  
& 39.2  & 34.2  & 20.7  & 22.0  & 25.7  & 36.4    & 21.3  & 23.2    & 45.3  & 27.9  & 23.3  & 42.5  & 28.1  & 30.0  \\
& Pixel2Mesh \cite{wang2018pixel2mesh}
& 51.5  & 40.7  & 43.4  & 50.1  & 40.2  & 55.9    & 29.1  & 52.3    & 50.9  & 60.0  & 31.2  & 69.4  & 40.1  & 47.3  \\
& 3DN \cite{wang20193dn}       
& 54.3  & 39.8  & 49.4  & 59.4  & 34.4  & 47.2    & 35.4  & 45.3    & 57.6  & 60.7  & 31.3  & 71.4  & 46.4  & 48.7  \\
& IMNET \cite{chen2019learning}    
& 55.4 & 49.5 & 51.5 & 74.5 & 52.2 & 56.2 & 29.6 & 52.6 &     52.3 &    64.1 &    45.0 &    70.9      & 56.6 & 54.6      \\
& 3DCNN \cite{xu2019disn}    
& 50.6 & 44.3 & 52.3 &    \textbf{76.9} &     52.6 &    51.5 &    36.2 &    58.0 &    50.5 &    67.2 &     50.3 &    70.9 & 57.4 & 55.3     \\
& OccNet \cite{mescheder2019occupancy}
& 54.7 & 45.2 & \textbf{73.2} &     73.1 &    50.2 &    47.9 &    \textbf{37.0} &     \textbf{65.3} &     45.8 &    67.1 &    \textbf{50.6} &     70.9 & 52.1 & 56.4     \\
& DISN \cite{xu2019disn}
& 57.5 & 52.9 &     52.3 &    74.3 &    54.3 &     56.4 &     34.7 &    54.9 &    59.2 &     65.9 &    47.9 &    72.9  & 55.9 & 56.9       \\
& Ours$_{cam}$
& \textbf{60.0}  & \textbf{53.4}   & 50.8  & 74.5  & \textbf{55.3}  & \textbf{57.8}    & 36.2  & 55.6    & \textbf{61.0}  & \textbf{68.5}  & 48.6  & \textbf{73.6}    & \textbf{61.3}  & \textbf{58.2} \\
& Ours
     & 66.6  & 60.3   & 56.4  & 80.2  & 64.7  & 63.7    & 48.5  & 61.5    & 71.9  & 73.5  & 58.1 & 78.1    & 68.8  & 65.6     \\
\hline
\end{tabular}
}
\end{center}
\end{table}

\begin{figure}
\centering
\includegraphics[height=3.4cm]{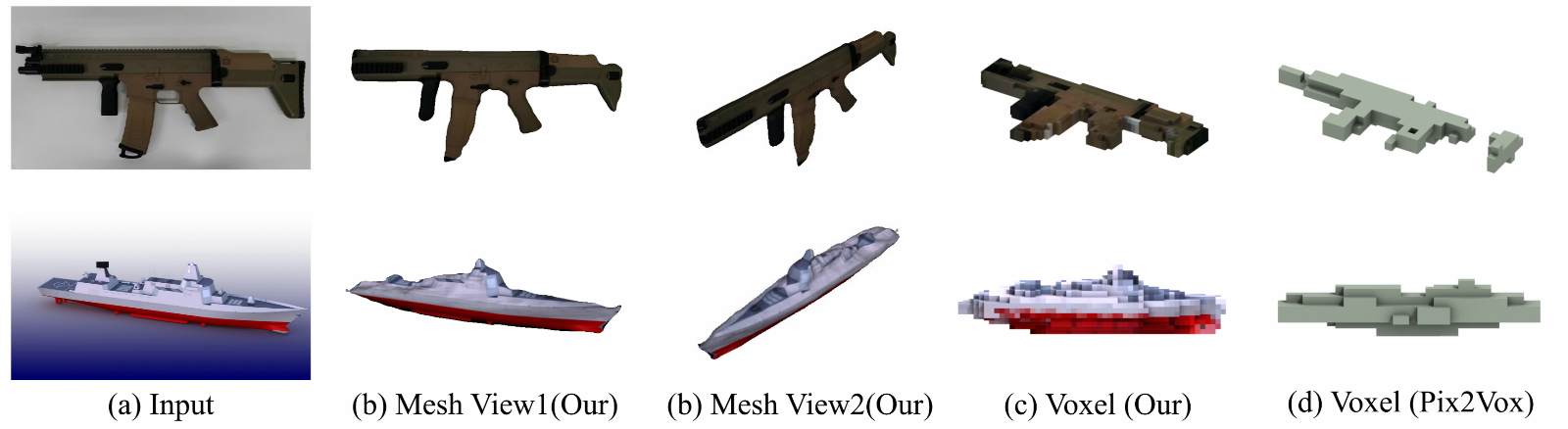}
\caption{Reconstruction results for online images. (a) indicates input images. (b) and (c) are our reconstruction results in mesh and voxel representation respectively. (d) shows the reconstruction results of Pixel2Vox \cite{xie2019pix2vox}. Ladybird naturally produces accurate uv-map for texturing.}
\label{fig:inthewild}
\end{figure}

Since ShapeNet is a synthesized dataset, we further provide quantitative evaluation on Pix3D \cite{sun2018pix3d} (Table~\ref{table:pix3d}), and some qualitative examples of in-the-wild images which are randomly selected from the internet (Figure~\ref{fig:inthewild}). These results show that Ladybird generalizes well to natural images. For the experiment on Pix3D, we fine-tune Ladybird and DISN \cite{xu2019disn} (both pre-trained on ShapeNet) on Pix3D train set, and use the ground truth camera poses and the segmentation masks.

\begin{table}[!t]
\begin{center}
\caption{Evaluations on Pix3D \cite{sun2018pix3d} test set. Metrics are CD ($\times 0.001$), and EMD ($\times 100$), computed on $2048$ points. Groud truth camera parameters are used.}
\label{table:pix3d}
\resizebox{\textwidth}{!}{
\begin{tabular}{c|c|ccccccccc|c}
Metric&  Method  & bed & bookcase & chair  & desk & misc & sofa & table  &  tool & wardrobe & Mean \\
\hline
CD
& DISN \cite{xu2019disn}
& 12.74  & 35.29  & 23.82  & 18.70  & 31.18  & 3.85    & 18.46 & 46.00   & \textbf{4.23}  & 18.51       \\
& Ours  & \textbf{5.73}  & \textbf{15.89}  & \textbf{13.03}  & \textbf{10.38}  & \textbf{30.34}    & \textbf{3.28} & \textbf{8.38}   & \textbf{28.39}  & 5.58   &  \textbf{10.02}   \\
\hline
EMD
& DISN \cite{xu2019disn}
& 2.84  & 4.65  & 3.97  & 4.04  & \textbf{4.53}  & 1.99    & 3.85 & 5.66   & 2.11  & 3.53        \\
& Ours  & \textbf{2.35}  & \textbf{3.07}  & \textbf{3.23}  & \textbf{2.77}  & 4.96    & \textbf{1.84} & \textbf{2.42}   & \textbf{3.68}  & \textbf{1.99}    & \textbf{2.75}   \\
\hline
IoU
& DISN \cite{xu2019disn}
& 71.2  & 43.0  & 59.0  & 53.7  & 48.8  & 89.4    & 57.8 & 37.3   & 85.6  & 64.4      \\
& Ours  & \textbf{78.2}  & \textbf{67.8}  & \textbf{66.5}  & \textbf{67.5}  & \textbf{49.5}    & \textbf{91.8} & \textbf{74.2}   & \textbf{58.4}  & \textbf{86.8}    &  \textbf{73.3}  \\
\hline
\end{tabular}
}
\end{center}
\end{table}

\section{Conclusion}

We study the impact of sample set discrepancy on the training efficiency of implicit field regression networks, and proposes to use FPS instead of Random sampling to select training points. We also propose to explore local feature fusion based on reflective symmetry to improve the reconstruction quality. Qualitatively and quantitatively we verify the efficiency of our methods through extensive experiments on large-scale dataset ShapeNet.

\section{Acknowledgement}
We would like to thank the anonymous reviewers for their helpful feedback and suggestions. We would like to thank Zilei Huang for his help in accelerating the data processing and debugging.

\section{Appendix}

\subsection{Validation accuracy}
In Table~\ref{table:val}, we report the SDF validation accuracy. The experimental setup is the same as that in Section 3.3, and our validation set consists of 6000 images. 
We see that Grid+FPS results in faster convergence and higher SDF validation accuracy.

\begin{table}[]
\begin{center}
\caption{Validation accuracy of different sampling method.}
\label{table:val}
\begin{tabular}{c|c|c|c|c|c|c}
\hline
Epoch &  1  & 2 & 3 & 5  & 10 & 30 \\
\hline
Grid+Random &  0.743 &	0.777 &	0.788 &	0.803 &	0.817 &	0.825 \\
\hline
Grid+FPS &  0.803 &	0.859 &	0.872 &	0.888 &	0.905 &	0.917 \\
\hline
\end{tabular}
\end{center}
\end{table}

\subsection{Spectrum, more on discrepancy}

\begin{figure}[]
\centering
\includegraphics[height=2.8cm]{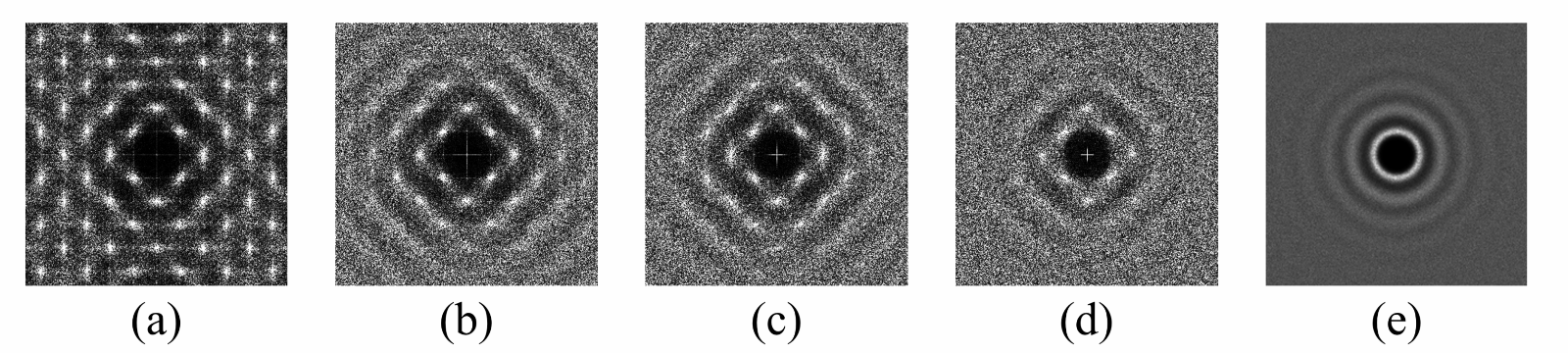}
\caption{Power spectra of (a) Grid+FPS, (b) Jitter+FPS ($\sigma=0.005$), (c) Jitter+FPS ($\sigma=0.01$), (d) Jitter+FPS ($\sigma=0.02$), (e) Blue noise.}
\label{fig:pow}
\end{figure}

FPS induces blue-noise behavior by construction. Gaussian Jitter+FPS gives a power spectrum with blue-noise characteristics (Figure~\ref{fig:pow}). However, Jitter+FPS gives higher discrepancy compared to Grid+FPS and worse 3D reconstruction results. Generating good 3D blue noise samples at $256^3$ resolution is computationally very expensive. Hence we excluded blue-noise samplers in this work.

The discrepancy depends on the initial sample size, final sample size, and their ratio. In Table~\ref{table:disp}, we report the Star Discrepancy (x0.01) of different samplers with varying initial sample size.
In the original FPS paper \cite{schlomer2011farthest}, the author gave a deterministic bounds on the distance between sample points (Theorem 4.2), which is used to prove that FPS is a uniform sampler. This analysis shields some lights on why FPS results in low-discrepancy, as it could lead to a deterministic bounds on discrepancy.

\begin{table}[!t]
\begin{center}
\caption{Mean ($\times 0.01$) and standard deviation ($\times 0.01$) of star discrepancy of different samplers. A+B means we first sample $n = 128^2, 256^2, 512^3$ points using sampling method A and then select a subset of size 2048 with method B.}
\label{table:disp}
\begin{tabular}{c|c|c|c|c|c}
\hline
Initial sample size & Metric & Grid+Random & Grid+FPS& Jitter+FPS &Sobol+FPS\\
\hline
$128\times128$ & Mean & 3.06 & 2.84 & 5.41 & 1.75 \\
     & Std & 0.34 & 0.18 & 0.16 & 0.08\\
\hline
$256\times256$ & Mean & 2.98 & 2.48 & 6.07 & 2.51 \\
     & Std & 0.26 & 0.16 & 0.77& 0.47 \\
\hline
$512\times512$ & Mean & 3.07 &  2.66 & 6.48 & 2.62 \\
     & Std & 0.5 & 0.1 &  0.31 & 0.23\\
\hline
\end{tabular}
\end{center}
\end{table}

\subsection{Marching Cube at higher resolution}
Using Ladybird configured as in Section 3.5, we run Marching Cube at different resolutions ($64^3$ and $512^3$). Due to the high memory and computation requirement at increased resolution, we only report CD for 100 objects that are randomly sampled from the ShapeNet test dataset. The results are summarized in Table~\ref{table:marchcube}.

\begin{table}
\begin{center}
\caption{Effect of Marching Cube resolution on the reconstruction results on 100 objects randomly sampled from ShapeNet test set. Metrics are class mean of CD ($\times 0.001$) computed on $2048$ points.}
\label{table:marchcube}
\begin{tabular}{c|c|c|c}
\hline
Resolution & Grid+Random & Grid+FPS &  Sobol+FPS \\
\hline
$64^3$  & 10.79  & 9.04 & 10.20\\
\hline
$512^3$ & 10.60  & 8.81 & 9.76\\
\hline
\end{tabular}
\end{center}
\end{table}

\subsection{Limitations}

The reconstruction quality of Ladybird is restricted by the input image resolution (currently 137x137). However, issues such as memory, speed and compatibility with pre-trained image networks need to be considered when increasing the input image resolution. We would like to address the problem of 3D reconstruction from a high resolution image in future work.

\begin{figure}[]
\centering
\includegraphics[height=5.0cm]{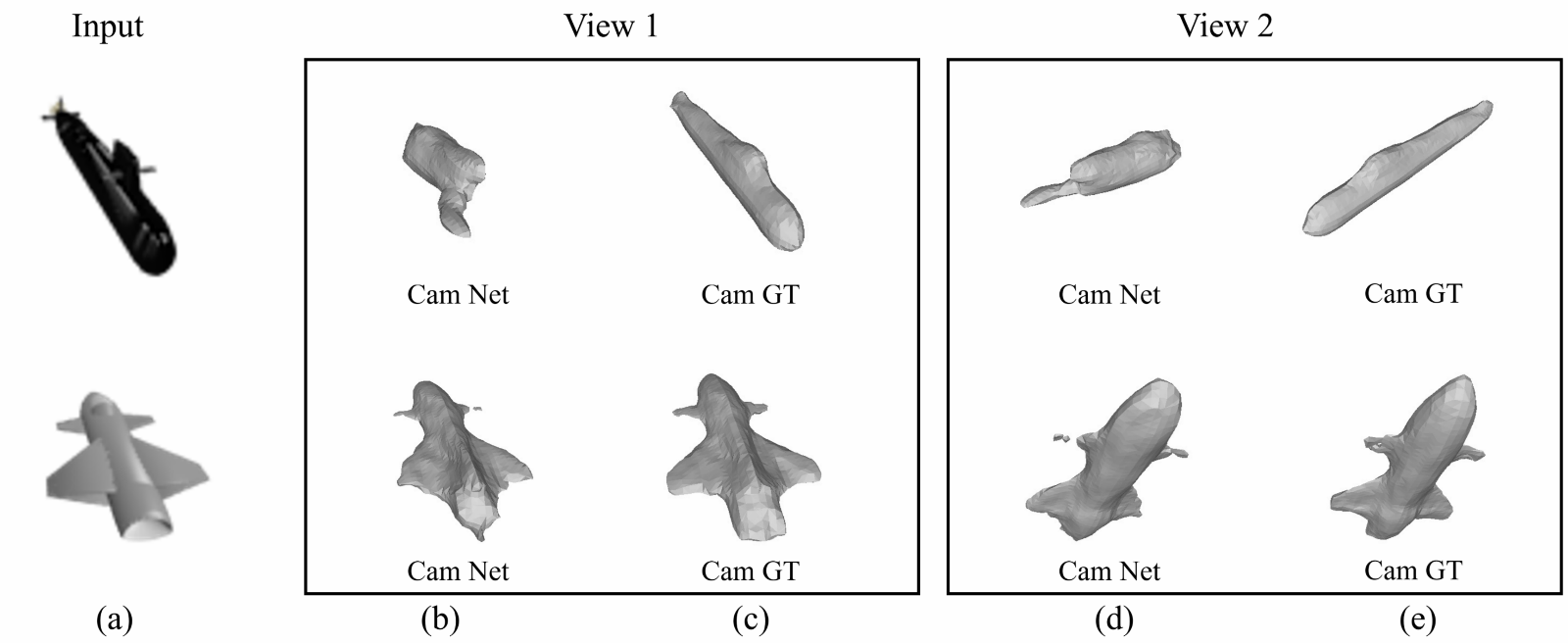}
\caption{Inaccurate estimation of camera pose leads to failures in reconstruction. (a) indicates the input images. (b) and (d) are the reconstruction results using estimated camera poses in two different views. (c) and (e) are the reconstruction results using ground truth camera poses in two different views.}
\label{fig:limit}
\end{figure}

Since we need to spatially align the image to the mesh and utilize the corresponding local features, accurate camera pose is crucial to our method (Figure~\ref{fig:limit}). A better camera pose estimation network will lead to significant improvement of our system.

\clearpage

%
%

\end{document}